\documentclass[10pt,twocolumn,letterpaper]{article}

\usepackage{authblk}
\usepackage{wacv}
\usepackage{times}
\usepackage{epsfig}
\usepackage{graphicx}
\usepackage{amsmath}
\usepackage{amssymb}
\usepackage{algorithm}
\usepackage{algorithmic}
\usepackage{subcaption}
\usepackage{xcolor}
\usepackage[pagebackref=true,breaklinks=true,letterpaper=true,colorlinks,bookmarks=false]{hyperref}

\wacvfinalcopy 

\def\wacvPaperID{607} 

\makeatletter
\newcommand{\printfnsymbol}[1]{%
  \textsuperscript{\@fnsymbol{#1}}%
}
\makeatother

\ifwacvfinal\fi
\begin{document}

\title{SieveNet: A Unified Framework for Robust Image-Based Virtual Try-On}
\author[2]{Surgan Jandial\thanks{equal contribution}\thanks{work done as part of Adobe MDSR internship program}}
\author[1]{Ayush Chopra\printfnsymbol{1}\thanks{corresponding author (ayuchopr@adobe.com)}}
\author[3]{Kumar Ayush\printfnsymbol{1}\thanks{work done while at Adobe}}
\author[1]{Mayur Hemani}
\author[4]{Abhijeet Kumar\printfnsymbol{2}}
\author[1]{Balaji Krishnamurthy}

\affil[1]{Media and Data Science Research Lab, Adobe}
\affil[2]{IIT Hyderabad}
\affil[3]{Stanford University}
\affil[4]{IIIT Hyderbad}

\maketitle
\ifwacvfinal\thispagestyle{empty}\fi

\begin{abstract}
Image-based virtual try-on for fashion has gained considerable attention recently. The task requires trying on a clothing item on a target model image. An efficient framework for this is composed of two stages: (1) warping (transforming) the try-on cloth to align with the pose and shape of the target model,  and  (2) a texture transfer module  to  seamlessly  integrate  the warped try-on cloth onto the target model image. Existing methods suffer from artifacts and distortions in their try-on output. In this work, we present SieveNet, a framework for robust image-based virtual try-on. Firstly, we introduce a multi-stage coarse-to-fine warping network to better model fine grained intricacies (while transforming the try-on cloth) and train it with a novel perceptual geometric matching loss. Next, we introduce a try-on cloth conditioned segmentation mask prior to improve the texture transfer network. Finally, we also introduce a duelling triplet loss strategy for training the texture translation network which further improves the quality of generated try-on result. We present extensive qualitative and quantitative evaluations of each component of the proposed pipeline and show significant performance improvements against the current state-of-the-art method. 
\end{abstract}

\section{Introduction}
Providing interactive shopping experiences is an important problem for online fashion commerce. Consequently, several recent efforts have been directed towards delivering smart, intuitive online experiences including clothing retrieval  \cite{deepfashion, gsn}, fine-grained tagging \cite{cvpr_attr, tagging_icip}, compatibility prediction \cite{contextaware, typeaware, tanmay2019augmented, hiranandani2017poster, ayushcontext} and virtual try-on \cite{CPVTON, VITON}. Virtual try-on is the visualization of fashion products in a personalized setting. The problem consists of trying on a specific garment on the image of a person. It is especially important for online fashion commerce because it compensates for the lack of a direct physical experience of in-store shopping.

\begin{figure}[t]
\begin{center}
  \includegraphics[width=0.7\linewidth]{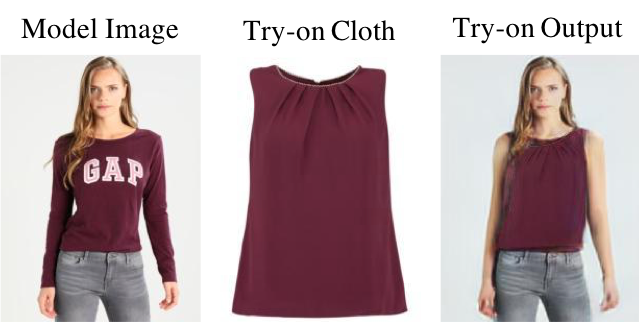}
\end{center}
  \caption{The task of image-based virtual try-on involves synthesizing a try-on output where the target model is wearing the try-on cloth while other characteristics of the model and cloth are preserved.}
\label{fig:tryonproblem}
\label{fig:long}
\label{fig:onecol}
\end{figure}

Recent methods based on deep neural networks \cite{CPVTON, VITON}, formulate the problem as that of conditional image generation. As depicted in Figure \ref{fig:tryonproblem}, the objective is to synthesize a new image (henceforth referred to as the try-on output) from two images - a try-on cloth and a target model image, such that in the try-on output the target model is wearing  the try-on cloth while the original body shape, pose and other model details (eg. bottom, face) are preserved.

\begin{figure}[t]
\begin{center}
  \includegraphics[width=0.9\linewidth]{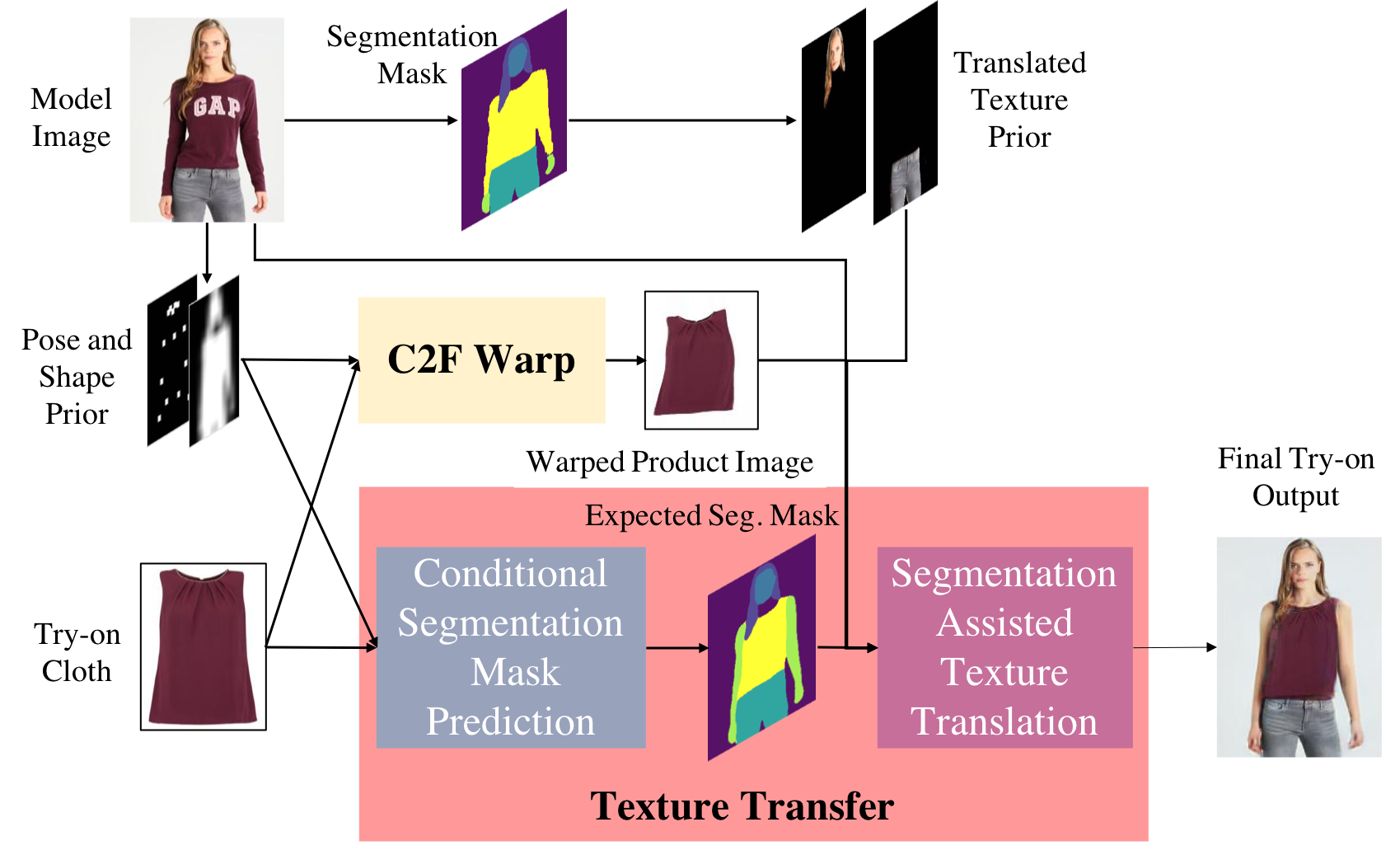}
  \caption{Inference Pipeline of the SieveNet framework}
  \label{fig:overview}
  \end{center}
\end{figure}

\begin{figure*}[t]
\begin{center}
  \includegraphics[width=0.9\linewidth]{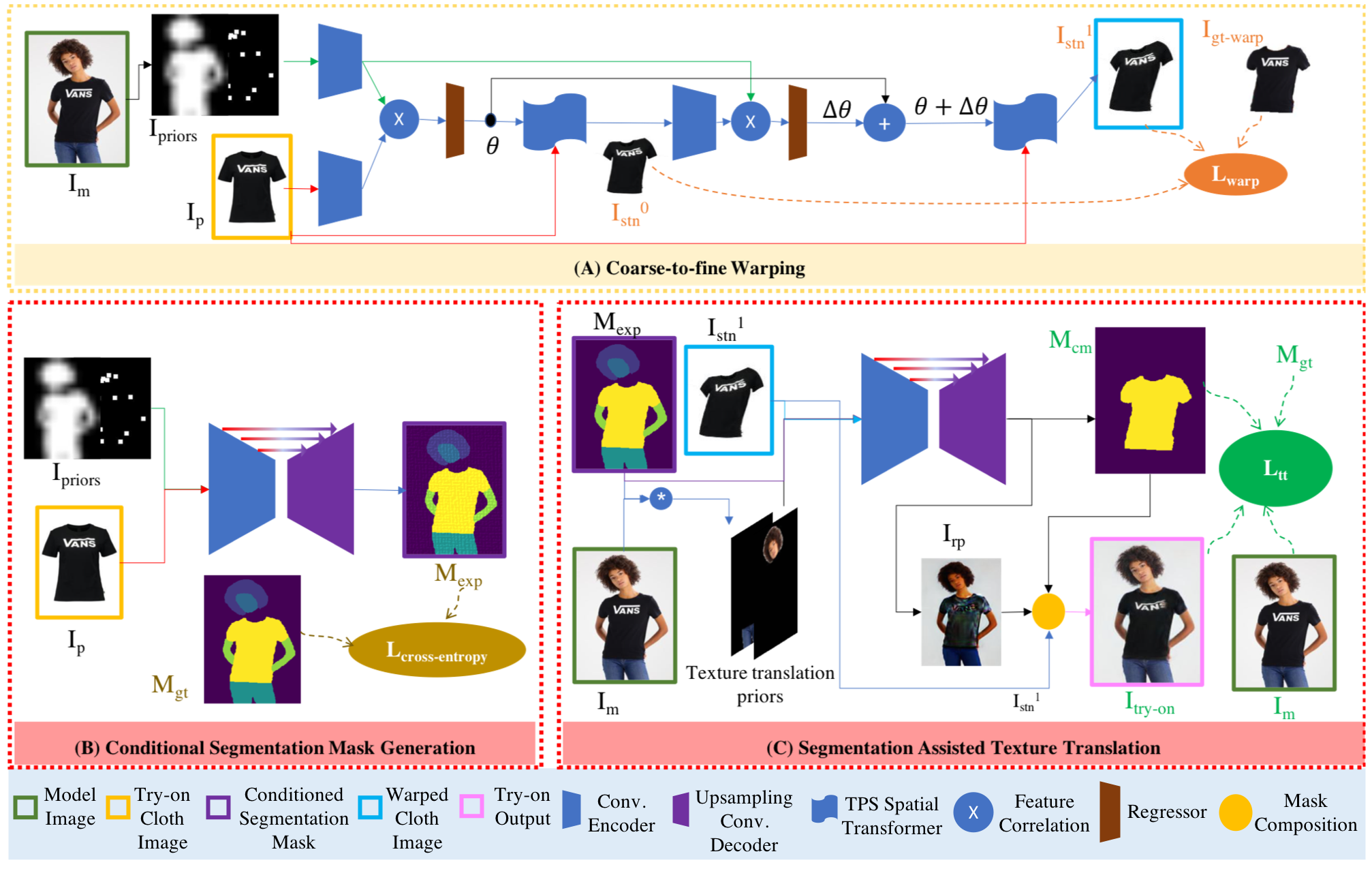}
  \caption{An overview of the training pipeline of SieveNet, containing (A) Coarse-to-Fine Warping Module, (B) Conditional Segmentation Mask Generation Module, and (C) Segmentation Assisted Texture Translation Module.}
  \label{fig:networks}
\end{center}
\end{figure*}

Successful virtual try-on experience depends upon synthesizing images free from artifacts arising from improper positioning or shaping of the try-on garment, and inefficient composition resulting in blurry or bleeding garment textures in the final try-on output. Current solutions \cite{VITON, CPVTON} suffer from these problems especially when the try-on cloth is subject to extreme deformations or when characteristics of the try-on cloth and original clothing item in target model differ. For example, transferring a half-sleeves shirt image to a target model originally in a full-sleeves shirt often results in texture bleeding and incorrect warping.
For alleviating these problems, we propose:
\begin{enumerate}
    \setlength\itemsep{0em}
    \item A multi-stage coarse-to-fine warping module trained with a novel perceptual geometric matching loss to better model fine intricacies while transforming the try-on cloth image to align with shape of the target model.
    \item A conditional segmentation mask generation module to assist in handling complexities arising from complex pose, occlusion and bleeding during the texture transfer process, and
    \item A duelling triplet loss strategy for training the texture translation network to further improve quality of the final try-on result.
\end{enumerate}
We show significant qualitative and quantitative improvement over the current state-of-the-art method for image-based virtual try-on. An overview of our SieveNet framework is presented in Figure \ref{fig:overview} and the training pipeline is detailed in Figure \ref{fig:networks}.

\section{Related Work}
Our work is related to existing methods for conditional person image synthesis that use pose and shape information to produce images of humans, and to existing virtual try-on methods - most notably \cite{CPVTON}.

\paragraph{Conditional Image Synthesis}
Ma et al. \cite{PoseGuided} proposed a framework for generating human images with pose guidance along with a refinement network trained using an adversarial loss. Deformable GANs \cite{DeformableGans} attempted to alleviate the misalignment problem between different poses by using an affine-transformation on the coarse rectangle region, and warped the parts on pixel-level. In \cite{vunet2018}, Esser et al. introduced a variational U-Net \cite{UNet} to synthesize the person image by restructuring the shape with stickman pose guidance. \cite{Pumarola2018UnsupervisedPI} applied CycleGAN directly to manipulate pose. However, all of these methods fail to preserve the texture details of the clothes in the output. Therefore, they cannot directly be applied to the virtual try-on problem.
\vspace{-8pt}
\paragraph{Virtual Try-On}
Initial works on virtual try-on were based on 3D modeling techniques and computer graphics. Sekine et al. \cite{singleshot3d} introduced a virtual fitting system that captures 3D measurements of body shape via depth images for adjusting 2D clothing images. 
Pons-Moll et al. \cite{clothCap} used a 3D scanner to automatically capture real clothing and estimate body shape and pose. Compared to graphics models, image-based generative models provide a more economical and computationally efficient solution. Jetchev et al. \cite{Jetchev_2017_ICCV} proposed a conditional analogy GAN to swap fashion articles between models without using person representations. They do not take pose variant into consideration, and during inference, they required the paired images of in-shop clothes and a wearer, which limits their applicability in practical scenarios. 
In \cite{VITON}, Han et al. introduced a virtual try-on network to transfer a desired garment on a person image. It uses an encoder-decoder with skip connections to produce a coarse clothing mask and a coarse rendered person image. It then uses a Thin-plate spline (TPS) based spatial transformation (from \cite{SpatialTransformers}) to align the garment image with the pose of the person, and finally a refinement stage to overlay the warped garment image on to the coarse person image to produce the final try-on image. Most recently Wang et al. \cite{CPVTON} present an improvement over \cite{VITON} by directly predicting the TPS parameters from the pose and shape information and the try-on cloth image. Both of these methods suffer from geometric misalignment, blurry and bleeding textures in cases where the target model is characterized by occlusion and where pose variation or garment shape variation is high. Our method, aligned to the approach in \cite{CPVTON}, improves upon all of these methods. SieveNet learns the TPS parameters in multiple stages to handle fine-grained shape intricacies and uses a conditional segmentation mask generation step to aid in handling of pose variation and occlusions, and improve textures. In Section \ref{comparisons}, we compare our results with \cite{CPVTON}.

\section{Proposed Methodology}
The overall process (Figure \ref{fig:overview}) comprises of two main stages - warping the try-on cloth to align with pose and shape of the target model, and transferring the texture from the warped output onto the target model to generate the final try-on image. We introduce three major refinements into this process. To capture fine details in the geometric warping stage, we use a two-stage spatial-transformer based warp module (Section \ref{coarseToFineWarping}). To prevent the garment textures from bleeding onto skin and other areas, we introduce a conditional segmentation mask generation module (Section \ref{condSegMask}) that computes an expected semantic segmentation mask to reflect the bounds of the target garment on the model, which in turn assists the texture translation network to produce realistic try-on results. We also propose two new loss computations - a perceptual geometric matching loss (Section \ref{perceptualGeometricLoss}) to improve the warping output, and a duelling triplet loss strategy (Section \ref{duellingLoss}) to improve the output from the texture translation network. 

\subsection{Inputs}
The framework uses the try-on cloth image ($I_p$), a 19-channel pose and body-shape map ($I_{priors}$) generated as described in \cite{CPVTON} as input to the various networks in our framework. $I_{priors}$ is a cloth-agnostic person representation created using the model image ($I_m$) to overcome the unavailability of ideal training triplets as discussed in \cite{CPVTON}.  A human parsing semantic segmentation mask ($M_{gt}$) is also used as ground-truth during training of the conditional segmentation mask generation module (described in Sections \ref{perceptualGeometricLoss} and \ref{condSegMask}). For training, the task is set such that data consists of paired examples where the model in $I_m$ is wearing the clothing product $I_p$.

\subsection{Coarse-to-Fine Warping}
\label{coarseToFineWarping}
The first stage of the framework warps the try-on product image ($I_p$) to align with the pose and shape of the target model ($I_m$). It uses the priors $I_{priors}$ as guidance for achieving this alignment. Warping is achieved using thin-plate spline (TPS) based spatial transformers \cite{SpatialTransformers}, as introduced in \cite{CPVTON} with a key difference that we learn the transformation parameters in a two-stage cascaded structure and use a novel perceptual geometric matching loss for training.

\subsubsection{Tackling Occlusion and Pose-variation}
We posit that accurate warping requires accounting for intricate modifications resulting from two major factors:
\begin{enumerate}
    \item Large variations in shape or pose between the try-on cloth image and the corresponding regions in the model image.
    \item Occlusions in the model image. For example, the long hair of a person may occlude part of the garment near the top.
\end{enumerate}
The warping module is formulated as a two-stage network to overcome these problems of occlusion and pose-variation. The first stage predicts a coarse-level transformation, and the second stage predicts the fine-level corrections on top of the coarse transformation. The transformation parameters from the coarse-level regression network ($\theta$) is used to warp the product image to produce an approximate warp output ($I_{stn}^0$). This output is then used to compute the fine-level transformation parameters ($\Delta\theta$) and the corresponding warp output ($I_{stn}^1$) is computed using ($\theta + \Delta\theta$) to warp the initial try-on cloth $I_p$ and not $I_{stn}^0$. This is done to avoid the artifacts from applying the interpolation in the spatial transformer twice. To facilitate the expected hierarchical behaviour, residual connections are introduced to offset the parameters of the fine-transformation with the coarse-transformation. 
The network structure is schematized in Figure \ref{fig:networks} (A). Ablation study to support the design of the network and losses is in Section \ref{ablateCoarseToFine}.

\subsection{Perceptual Geometric Matching Loss}
\label{perceptualGeometricLoss}
The interim ($I_{stn}^0$) and final ($I_{stn}^1$) output from the warping stage are subject to a matching loss $L_{warp}$ against the $I_{gt-warp}$ (segmented out from the model image) during training. $L_{warp}$ is defined below which includes a novel perceptual geometric matching loss $L_{pgm}$ component.
The intuition behind this loss component $L_{pgm}$ is to have the second stage warping incrementally improve upon that from the first stage.
\begin{equation}
\begin{gathered}
L_{warp} = \lambda_1L_s^0 + \lambda_2L_s^1 + \lambda_3L_{pgm}\\
L_s^0 = |I_{gt-warp} - I_{stn}^0| \\
L_s^1 = |I_{gt-warp} - I_{stn}^1|
\end{gathered}
\end{equation}
Here, $I_{gt-warp} = I_m * M_{gt}^{cloth}$, and $L_{pgm}$ is the perceptual geometric matching loss which comprises of two components. $I_{gt-warp}$ is the cloth worn on the target model in $I_m$ and $M_{gt}^{cloth}$ is the binary mask representing the cloth worn on the target model.
\begin{equation}
L_{pgm} =  \lambda_4L_{push} + \lambda_5L_{align}
\end{equation}
Minimizing $L_{push}$ pushes the second stage output $I_{stn}^1$  closer to the ground-truth $I_{gt-warp}$ compared to the first stage output.
\begin{equation}
L_{push} =  k * L_s^1 - |I_{stn}^1 - I_{stn}^0|\\
\end{equation}

The scalar $k$ is a multiplicative margin used to ensure stricter bound for the difference ($k=3$ is used for our experiments).
\begin{figure}[t]
\begin{center}
  \includegraphics[width=0.7\linewidth]{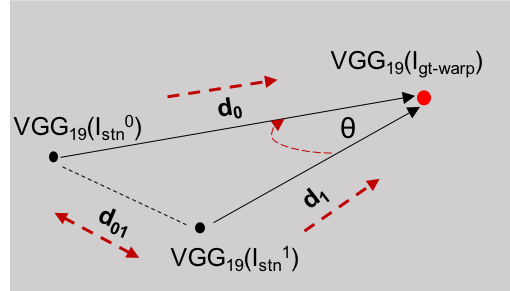}
  \caption{Visualization of the Perceptual Geometric Matching Loss in VGG-19 Feature Space.}
  \end{center}
\label{fig:pgm}
\end{figure}

For $L_{align}$, $I_{stn}^0$, $I_{stn}^1$ and $I_{gt-warp}$ are first mapped to the VGG-19 activation space, and then the loss attempts to align the difference vectors between $I_{stn}^0$ and $I_{gt-warp}$, and $I_{stn}^1$ and $I_{gt-warp}$ in the feature space. 
\begin{equation}
\begin{gathered}
V^0 = VGG(I_{stn}^0) - VGG(I_{gt-warp})\\
V^1 = VGG(I_{stn}^1) - VGG(I_{gt-warp})\\
L_{align} =  (CosineSimilarity(V^0, V^1) - 1)^2
\end{gathered}
\end{equation}

Minimizing $L_{align}$ facilitates the goal of minimizing $L_{push}$.

\subsection{Texture Transfer}
Once the product image is warped to align with the pose and shape of the target model, the next stage transfers the warped product to the model image. This stage computes a rendered model image, and a fractional composition mask to compose the warped product image onto the rendered model image. We break down this stage into two steps - conditional segmentation mask prediction and segmentation assisted texture translation.

\begin{figure}[t]
\begin{center}
\includegraphics[width=0.6\linewidth]{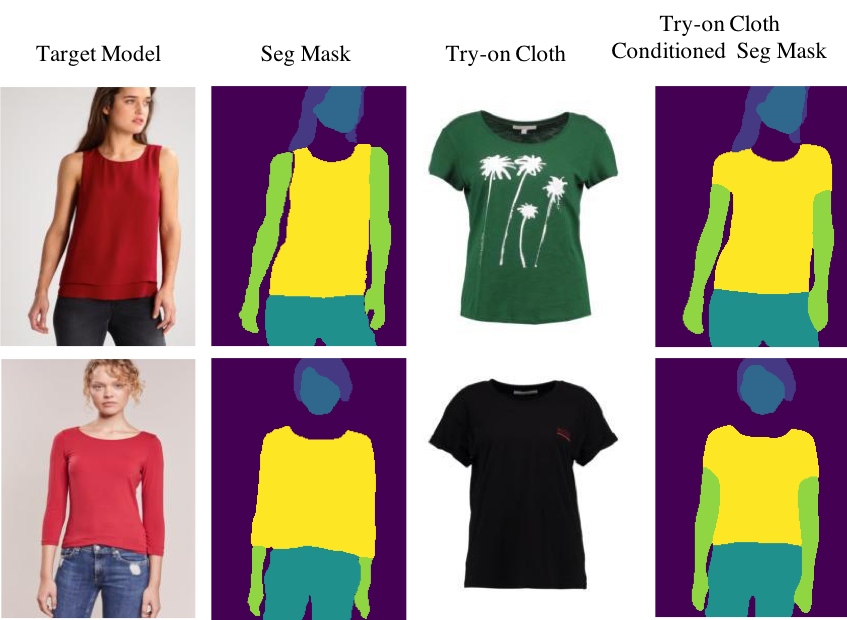}
\end{center}
\caption{Illustrating work of Conditional Segmentation Mask Prediction Network}
\label{fig:demoSegCorrect}
\end{figure}

\subsubsection{Conditional Segmentation Mask Prediction}
\label{condSegMask}
A key problem with existing methods is their inability to accurately honor the bounds of the clothing product and human skin. The product pixels often bleeds into the skin pixels (or vice-versa), and in the case of self-occlusion (such as with the case of folded arms), the skins pixels may get replaced entirely. This problem is exacerbated for cases where the try-on clothing item has a significantly different shape than the clothing in the model image. Yet another scenario that aggravates this problem is when the target model is in a complex pose. To help mitigate these problems of bleeding and self-occlusion as well as to handle variable and complex poses, we introduce a conditional segmentation mask prediction network.

Figure \ref{fig:networks} (B) illustrates the schematics of the network. It takes the pose and shape priors ($I_{priors}$) and the product image ($I_p$) as input, to generate an ``expected'' segmentation mask ($M_{exp}$). This try-on clothing conditioned seg. mask represents the expected segmentation of the generated try-on output where the target model is now wearing the try-on cloth. Since we are constrained to train with coupled data ($I_p$ and $I_m$), this expected (generated) segmentation mask ($M_{exp}$) is matched against the ground-truth segmentation mask ($M_{gt}$) itself. We intend to highlight that the network is able to generalize to unseen models at inference since it learns from a sparse clothing agnostic input ($I_{priors}$) that does not include any effects of worn cloth in target model image or segmentation mask (to avoid learning identity). At inference time, the generated $M_{exp}$ is directly used downstream. Figure \ref{fig:demoSegCorrect} demonstrates some examples of the corrected segmentation masks generated with our network, and ablation studies to support the use of the conditional segmentation mask is in Section \ref{ablateCondSegMask}.

The network (a 12-layer U-Net \cite{UNet} like architecture) is trained with a weighted cross-entropy loss, which is the standard cross-entropy loss for semantic segmentation with increased weights for skin and background classes. The weight of the skin is increased to better handle occlusion cases, and the background weight is increased to stem bleeding of the skin pixels into the background.

\subsubsection{Segmentation Assisted Texture Translation}
The last stage of the framework uses the expected segmentation mask ($M_{exp}$), the warped product image ($I_{stn}^1$), and unaffected regions from the model image ($I_m$) to produce the final try-on image.
The network is a 12-layer U-Net \cite{UNet} that takes the following inputs:
\begin{itemize}
    \setlength\itemsep{0em}
    \item The warped product image $I_{stn}^1$
    \item The expected seg. mask $M_{exp}$, and
    \item Pixels of $I_m$ for the unaffected regions, (Texture Translation Priors in Figure \ref{fig:networks}). E.g. face and bottom cloth, if a top garment is being tried-on.
\end{itemize}
The network produces two output images - an RGB rendered person image ($I_{rp}$) and a composition mask $M_{cm}$, which are combined with the warped product image $I_{stn}^1$ using the following equation to produce the final try-on image:
\begin{equation}
    I_{try-on} = M_{cm} * I_{stn}^1 + (1 - M_{cm}) * I_{rp}
\end{equation}

Because the unaffected parts of the model image are provided as prior, the proposed framework is also able to better translate texture of auxiliary products such as bottoms onto the final generated try-on image (unlike in \cite{CPVTON} and \cite{VITON}).

The output of the network is subject to the following matching losses based on $L_1$ distance and a perceptual distance based on VGG-19 activations:
\begin{equation}
\begin{gathered}
    L_{tt} = L_{l1} + L_{percep} + L_{mask}\\
    L_{l1} = |I_{try-on} - I_m|\\
    L_{percep} = |VGG(I_{try-on}) - VGG(I_m)|\\
    L_{mask} = |M_{cm} - M_{gt}^{cloth}|
\end{gathered}
\end{equation}
The training happens in multiple phases. The first $K$ steps of training is a conditioning phase that minimizes the $L_{tt}$ to produce reasonable results. The subsequent phases (each lasting $T$ steps) employ the $L_{tt}$ loss augmented with a triplet loss (Section \ref{duellingLoss}) to fine-tune the results further. This strategy further improves the output significantly (see ablation study in Section \ref{ablateduellingLoss}).

\subsubsection{Duelling Triplet Loss Strategy}
\label{duellingLoss}
A triplet loss is characterized by an anchor, a positive and a negative (w.r.t the anchor), with the objective being to simultaneously push the anchor result towards the positive and away from the negative. In the duelling triplet loss strategy, we pit the output obtained from the network with the current weights (anchor) against that from the network with weights from the previous phase (negative), and push it towards the ground-truth (positive). As training progresses, this online hard negative mining strategy helps push the results closer to the ground-truth by updating the negative at discrete step intervals ($T$ steps). In the fine-tuning phase, at step $i$ ($i > K$) the triplet loss is computed as:
\begin{equation}
\begin{gathered}
i_{prev} = K + T*(\lfloor\frac{i-K}{T}\rfloor - 1) \\
D_{neg}^i = |I_{try-on}^i - I_{try-on}^{i_{prev}}|\\
D_{pos}^i = |I_{try-on}^i - I_m|\\
L_{d}^i = \max(D_{pos}^i - D_{neg}^i, 0)
\end{gathered}
\end{equation}
Here $I_{try-on}^i$ is the try-on image output obtained from the network with weights at the $i^{th}$ iteration.
The overall loss with the duelling triplet strategy in use is then computed for a training step $i$ as:
\begin{equation}
L_{tryon}^i = \begin{cases}
L_{tt} & i \leq K\\
L_{tt} + L_{d}^i & i \textgreater K
\end{cases}
\end{equation}

\begin{figure}[t]
\begin{center}
  \includegraphics[width=0.85\linewidth]{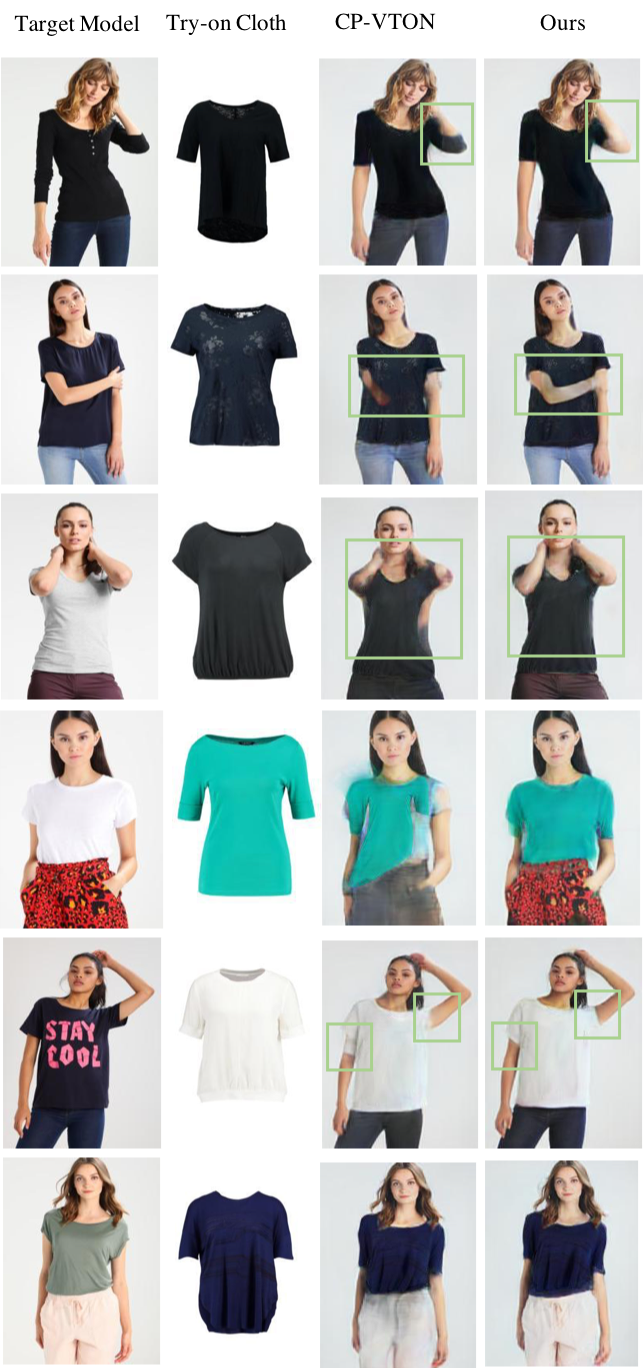}
\end{center}
\caption{SieveNet can generate more realistic try-on results compared to the state-of-the-art CP-VTON.}
\label{fig:versuscpvton}
\end{figure}

\section{Experiments}
\subsection{Datasets}
We use the dataset collected by Han et al. \cite{VITON} for training and testing. It contains around 19,000 images of front-facing female models and the corresponding upper-clothing isolated product images. There are 16253 cleaned pairs, which are split into a training set and a testing set with 14221 and 2032 pairs, respectively. The images in the testing set are rearranged into unpaired sets for qualitative evaluation and kept paired for quantitative evaluation otherwise. 
\subsection{Implementation Details}
All experiments are conducted on 4 NVIDIA 1080Ti on a machine with 16 GB RAM. The hyper-parameter configurations were as follows: batch size=$16$, epochs=$15$, optimizer=Adam\cite{kingma2014adam}, lr=$0.002$, $\lambda_1$=$\lambda_2$=$\lambda_3$=$1$, $\lambda_4$=$\lambda_5$=$0.5$.

\subsection{Quantitative Metrics}
To effectively compare the proposed approach against the current state-of-the-art, we report our performance using various metrics including Structural Similarity (SSIM) \cite{SSIM}, Multiscale-SSIM (MS-SSIM) \cite{MSSIM}, Fr\'echet Inception Distance (FID) \cite{FrechetInceptionDistance}, Peak Signal to Noise Ratio (PSNR), and Inception Score (IS) \cite{InceptionScore}. We adapt the Inception Score metric in our case as a measure of generated image quality by estimating similarity of generated image distribution to the ground truth distribution. For computing pairwise MS-SSIM and SSIM metrics, we use the paired test data.

\subsection{Baselines}
CP-VTON\cite{CPVTON} and VITON \cite{VITON} are the latest image based virtual try-on methods, with CP-VTON being the current state-of-the-art.  In particular, \cite{VITON} directly applied shape context \cite{shapecontext} matching to compute the transformation mapping. By contrast, \cite{CPVTON} estimates the transformation mapping using a convolutional network and has superior performance than \cite{VITON}. We therefore use results from CP-VTON \cite{CPVTON} as our baseline.

\section{Results}
The task of virtual try-on can be broadly broken down into two stages, \textit{warping} of the product image and \textit{texture transfer} of the warped product image onto the target model image. We conduct extensive quantitative and qualitative evaluations for both stages to validate the effectiveness of our contributions (coarse-to-fine warping trained with perceptual geometric matching loss, try-on cloth conditioned segmentation mask prior, and the duelling triplet loss strategy for training the texture translation network) over the existing baseline CP-VTON \cite{CPVTON}. 

\subsection{Quantitative Results}
Table \ref{table:quant} summarizes the performance of our proposed framework against CP-VTON on benchmark metrics for image quality (IS, FID and PSNR) and pair-wise structural similarity (SSIM and MS-SSIM). To highlight the benefit of our contributions in warp and texture transfer, we experiment with different warping and texture transfer configurations (combining modules from CP-VTON with our modules).
All scores progressively improve as we swap-in our modules. Using our final configuration of coarse-to-fine warp (C2F) and segmentation assisted texture translation with duelling triplet strategy (SATT-D) improved FID from 20.331 (for CP-VTON) to 14.65. Also, PSNR increased by around 17\% from 14.554 to 16.98. 
While a higher Inception score (IS) is not necessarily representative of output quality for virtual try-on, we argue that the proposed approach is able to better model the ground truth distribution as it produces an IS  (2.82 $\pm$ 0.09) which is closer to the IS for ground-truth images in the test set (2.83 $\pm$ 0.07) than CP-VTON (2.66 $\pm$ 0.14).
These quantitative claims are further substantiated in subsequent sections where we qualitatively highlight the benefit from each of the components.
\begin{table*}[]
\begin{center}
\begin{tabular}{|l|l|l|l|l|l|}
\hline
Configuration & SSIM  & MS-SSIM & FID    & PSNR   & IS           \\
\hline
GMM + TOM (CP-VTON)   & 0.698 & 0.746  & 20.331 & 14.544 & 2.66 $\pm$ 0.14   \\
GMM + SATT   & 0.751 & 0.787  & 15.89  & 16.05  & 2.84 $\pm$ 0.13 \\
C2F + SATT   & 0.755 & 0.794  & 14.79  & 16.39  & 2.80 $\pm$ 0.08 \\
\textbf{C2F + SATT-D (SieveNet)} & \textbf{0.766} & \textbf{0.809}  & \textbf{14.65}  & \textbf{16.98}  & \textbf{2.82 $\pm$ 0.09} \\
\hline
\end{tabular}
\caption{Quantitative comparison of Proposed vs CP-VTON. GMM, TOM are the warping and texture transfer modules from CP-VTON. C2F is the coarse-to-fine warp network and SATT is the segmentation assisted texture translation network we introduce in this framework. SATT-D is SATT trained with the duelling triplet loss strategy.}
\label{table:quant}
\end{center}
\end{table*}

\subsection{Qualitative Results}
\label{comparisons}
Figure \ref{fig:versuscpvton} presents a comparison of results of the proposed framework with those of CP-VTON. The results are presented to compare the impact on different aspects of quality - skin generation (row 1), handling occlusion (row 2), variation in poses (row 3), avoiding bleeding (row 5), preserving unaffected regions (row 4), better geometric warping (row 4) and overall image quality (row 6). For all aspects, our method produces better results than CP-VTON for most of the test images. These observations are corroborated by the quantitative results reported in Table \ref{table:quant}. 

\subsection{Ablation Studies}
In this section, we present a series of ablation studies to qualitatively highlight the particular impact of each our contributions: the coarse-to-fine warp, try-on product conditioned segmentation prediction and the duelling triplet loss strategy for training the texture translation module. 

\begin{figure}[t]
\begin{center}
  \includegraphics[width=0.8\linewidth]{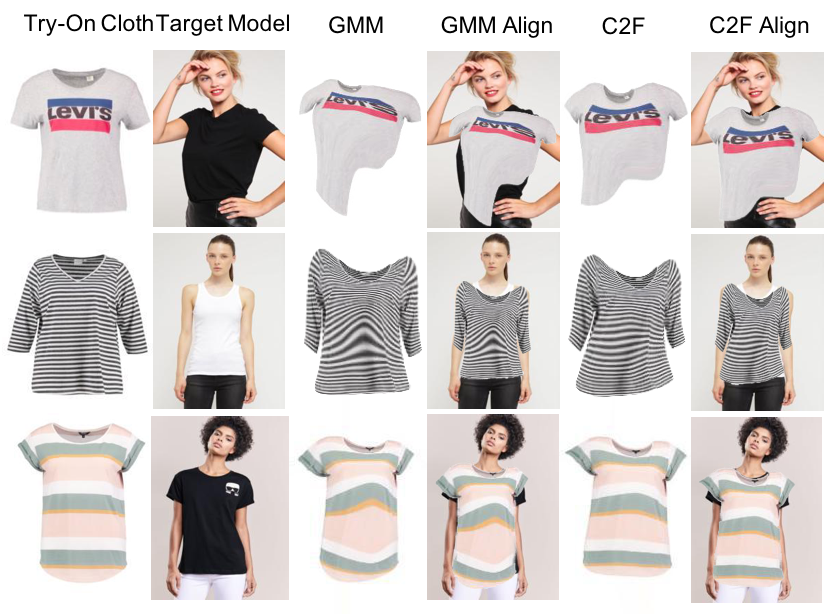}
\caption{Comparison of our C2F warp results with GMM warp results. Warped clothes are directly overlaid onto target persons for visual checking. C2F produces robust warp results which can be seen from preservation of text (row 1) and horizontal stripes (row 2, row 3) along with better fitting. GMM produces highly unnatural results.}
\label{fig:c2fablation}
\end{center}
\end{figure}
\vspace{-6 pt}
\paragraph{Impact of Coarse-to-Fine Warp}
\label{ablateCoarseToFine}
\begin{figure}[t]
\begin{center}
  \includegraphics[width=0.8\linewidth]{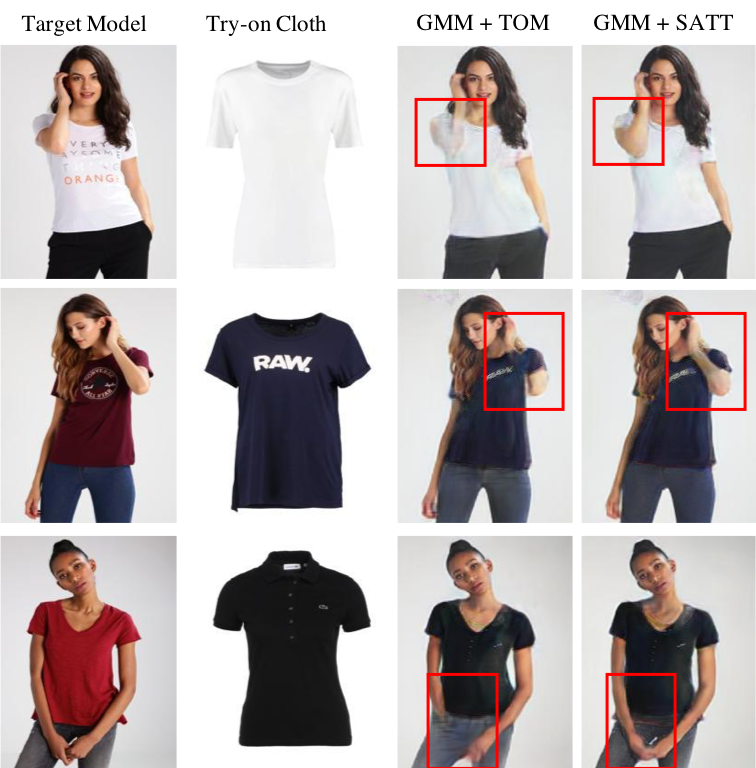}
\end{center}
\caption{Using the conditional segmentation mask as prior to texture transfer aids in better handling of complex pose, occlusion and helps avoid bleeding.}
\label{fig:condsegmask}
\end{figure}

\begin{figure}[t]
\begin{center}
  \includegraphics[width=0.8\linewidth]{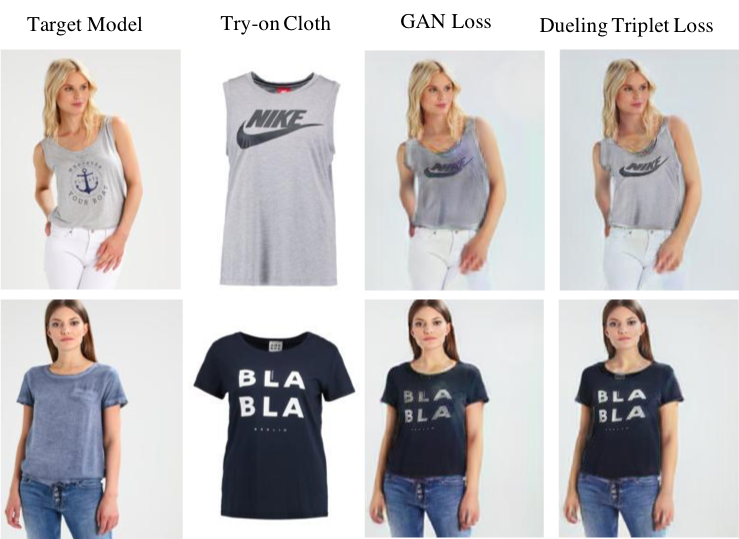}
\end{center}
\caption{Proposed Duelling Triplet Loss helps in better handling of texture and avoiding blurry effects in generated results than the GAN Loss.}
\label{fig:long}
\label{fig:ganablate}
\end{figure}

Figure \ref{fig:c2fablation} presents sample results comparing outputs of the proposed coarse-to-fine warp approach against the geometric matching module (GMM) used in CP-VTON \cite{CPVTON}.
Learning warp parameters in a multi-stage framework helps in better handling of large variations in model pose and body-shape in comparison to the single stage warp in \cite{CPVTON}. The coarse-to-fine (C2F) warp module trained with our proposed perceptual geometric matching loss does a better job at preserving textures and patterns on warping.
This is further corroborated through the quantitative results in Table \ref{table:quant} (row 2 vs row 3).
\begin{figure}[t]
\begin{center}
  \includegraphics[width=0.8\linewidth]{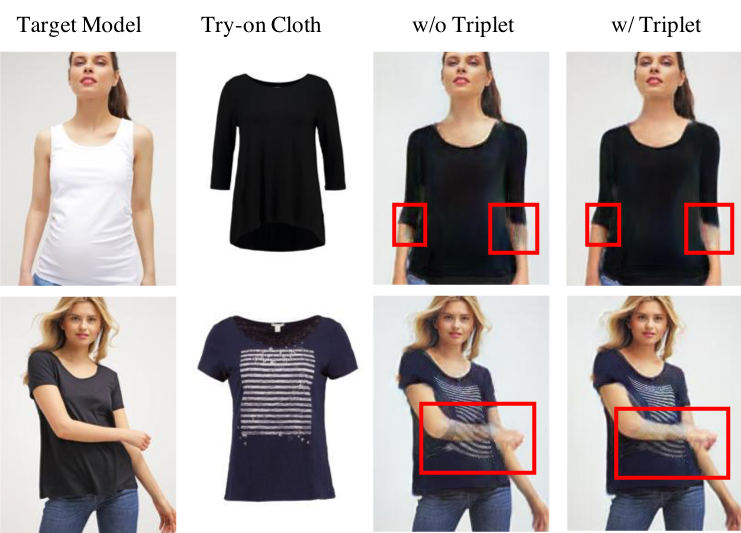}
\caption{Finetuning texture translation with the duelling triplet strategy refines quality of generated images by handling occlusion and avoiding bleeding.}
\label{fig:duelling}\end{center}
\end{figure}
\vspace{-6 pt}
\paragraph{Impact of Duelling Triplet Loss}
\label{ablateduellingLoss}
In Figure \ref{fig:duelling}, we present sample results depicting the particular benefit of training the texture translation network with the duelling triplet strategy. As highlighted by the results, using the proposed triplet loss for online hard negative mining in the fine-tuning stage refines the quality of the generated results. This arises from better handling of occlusion, bleeding and skin generation. These observations are corroborated by results in Table \ref{table:quant} (row 3 vs 4).
\vspace{-8 pt}
\paragraph{Impact of Conditional Segmentation Mask Prediction}
\label{ablateCondSegMask}
Figure \ref{fig:condsegmask} presents results obtained by training the texture transfer module of CP-VTON (TOM) \cite{CPVTON} with an additional prior of the try-on cloth conditioned segmentation mask. It can be observed that this improves handling of skin generation, bleeding and complexity of poses. Providing the expected segmentation mask of the try-on output image as prior equips the generation process to better handle these issues. These observations are corroborated through results in Table \ref{table:quant} (row 1 vs 2).
\vspace{-8 pt}
\paragraph{Impact of Adversarial Loss on Texture Transfer}
Many recent works on conditional image generation \cite{CPVTON, SwapNet, PoseGuided} employ a discriminator network to help improve quality of generated results. However, we observe that fine-tuning with the duelling triplet strategy instead results in better handling of texture and blurring in generated images without the need for any additional trainable parameters. Sample results in Figure \ref{fig:ganablate} corroborate the claim.
\begin{figure}[t]
\begin{subfigure}{0.5\textwidth}
  \centering
  \includegraphics[width=0.85\linewidth]{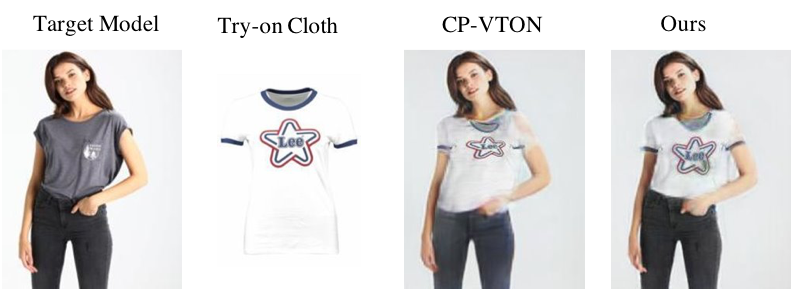}
  \caption{Failure in correctly occluding the back portion of the t-shirt.} 
\end{subfigure}
\newline
\begin{subfigure}{0.5\textwidth}
\begin{center}
  \centering
  \includegraphics[width=0.85\linewidth]{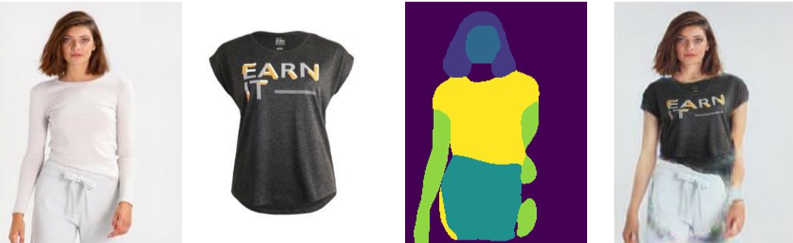}
  \caption{Failure in predicting the correct segmentation mask owing to errors in key-point prediction.} 
\end{center}
\end{subfigure}
\caption{Failure Cases}
\label{fig:limitations}
\end{figure}
\vspace{-8 pt}
\paragraph{Failure Cases}
While SieveNet performs significantly better than existing methods, it has certain limitations too. Figure \ref{fig:limitations} highlights some specific failure cases. In some cases, generated result is unnatural due to presence of certain artifacts (as the gray neckline of the t-shirt in the example in row 1) that appear in the output despite the correct fit and texture being achieved. This problem can be alleviated if localized fine-grained key-points are available. Further, texture quality in try-on output may be affected by errors in the predicted conditional segmentation mask. This happens due to errors in predicting pose key-points. For instance, this may happen in model images with low-contrast regions (example in row 2). Using dense pose information or a pose prediction network can help alleviate this problem.

\section{Conclusion}
In this work, we propose SieveNet, a fully learnable image-based virtual try-on framework. We introduce a coarse-to-fine cloth warping network trained with a novel perceptual geometric matching loss to better model fine-grained intricacies while transforming the try-on cloth image to align with shape of the target model. Next, we achieve accurate texture transfer using a try-on cloth conditioned segmentation mask prior and training the texture translation network with a novel duelling triplet loss strategy. We report qualitatively and quantitatively superior results over the state-of-the-art methods.


{\small
\bibliographystyle{ieee}
\bibliography{main}
}

\end{document}